\documentclass{article} 
\usepackage{iclr2026_conference,times}
\usepackage{makecell}
\usepackage{multirow}

\usepackage{amsmath,amsfonts,bm}

\def\eqref#1{equation~\ref{#1}}

\def\1{\bm{1}}

\DeclareMathAlphabet{\mathsfit}{\encodingdefault}{\sfdefault}{m}{sl}
\SetMathAlphabet{\mathsfit}{bold}{\encodingdefault}{\sfdefault}{bx}{n}

\usepackage{hyperref}
\usepackage{url}
\usepackage{booktabs}
\usepackage{array}
\usepackage{tabularx}
\usepackage{xurl}
\usepackage{algorithm}
\usepackage{algpseudocode}
\usepackage{graphicx}
\usepackage{wrapfig}
\usepackage{capt-of}
\usepackage{xcolor} 
\usepackage{soul} 
\usepackage{amssymb} 
\usepackage{xspace} 
\usepackage{adjustbox} 
\usepackage{subcaption} 

\newcommand{\chan}[1]{{\color{black} #1}} 
\newboolean{showcomments}
\setboolean{showcomments}{false} 
\ifthenelse{\boolean{showcomments}}
 { \newcommand{\mynote}[2]{
      \fbox{\bfseries\sffamily\scriptsize#1}
        {\small$\blacktriangleright${{{#2}\bf }}$\blacktriangleleft$}}}
        { \newcommand{\mynote}[2]{}}

\newcommand{\ak}[1]{\mynote{Ahmed}{\hl{#1}}}

\newcommand{\VDA}{$\hat{\text{A}}_{12}$\xspace}

\title{Towards a more efficient bias detection\\in financial language  models}

\author{
{Firas Hadj Kacem$^1$}, 
{Ahmed Khanfir$^{2,1}$} and 
{Mike Papadakis$^1$}\\
{$^1$ SnT, University of Luxembourg, Luxembourg}\\
{$^2$ RIADI, ENSI, University of Manouba, Tunisia}\\
\texttt{firashadjkacem@ieee.org},\\ 
\texttt{ahmed.khanfir@ensi-uma.tn} and \\
\texttt{michail.papadakis@uni.lu}
}

\iclrfinalcopy 
\begin{document}

\maketitle

\begin{abstract}
Bias in financial language models constitutes a major obstacle to their adoption in real-world applications. Detecting such bias is challenging, as it requires identifying inputs whose predictions change when varying properties unrelated to the decision, such as demographic attributes. Existing approaches typically rely on exhaustive mutation and pairwise prediction analysis over large corpora, which is effective but computationally expensive—particularly for large language models—and can become impractical in continuous retraining and releasing processes. Aiming at reducing this cost, we conduct a large-scale study of bias in five financial language models, examining similarities in their bias tendencies across protected attributes and exploring cross-model–guided bias detection to identify bias-revealing inputs earlier. 
Our study uses approximately \chan{17k} real financial news sentences, mutated to construct over \chan{125k} original–mutant pairs. Results show that all models exhibit bias under both atomic \chan{(0.58\%–6.05\%)} and intersectional \chan{(0.75\%–5.97\%)} settings. Moreover, we observe consistent patterns in bias-revealing inputs across models, enabling substantial reuse and cost reduction in bias detection. For example, up to \chan{73\%} of FinMA’s biased behaviors can be uncovered using only \chan{20\%} of the input pairs when guided by properties derived from DistilRoBERTa outputs.

\end{abstract}

\section{Introduction}
\label{sec:introduction}
The rapid progress in artificial intelligence has led to increasing interest in language models for tasks such as financial news analysis, risk assessment, and decision support~\citep{xie2023pixiu}. Recent advances in transformer-based architectures enabled the development of specialized financial language models outperforming general-purpose ones on financial tasks~\citep{yang2023fingpt, xie2023pixiu, araci2019finbert}. However, their adoption in real-world systems remains limited. 

Among the primary barriers to adoption is the presence of bias in language models~\citep{guo2024biasLLMs}, as biased predictions can result in discriminatory outcomes affecting individuals or groups. In the financial domain, such risks are amplified by strict regulatory requirements. Thus, bias detection in financial language models has become a pressing concern for researchers and practitioners. \\
Existing studies have shown that language models may exhibit bias toward protected attributes such as gender, race, or physical features~\citep{kiritchenko2018examining, bolukbasi2016man}. Nevertheless, most prior work focuses on general-purpose language models or evaluates bias using small datasets~\citep{asyrofi2021biasfinder}. Moreover, bias detection techniques typically rely on exhaustive testing strategies; costly and difficult to scale to multiple models, continuous releases, or large corpora. 
Thus, there is a lack of empirical evidence on whether different financial language models exhibit similar bias patterns and if bias-revealing inputs can be efficiently identified and reused across models. 

This work bridges these gaps by conducting a large-scale empirical study of bias in financial language models, i.e., two generative LLMs (FinMA~\citep{yang2023fingpt} and FinGPT~\citep{yang2023fingpt}) and three encoder-based models (FinBERT~\citep{araci2019finbert} and fine-tuned versions of DeBERTa-v3~\citep{romero2024deberta} and DistilRoBERTa~\citep{romero2024distilroberta}), on a dataset of real financial sentences, systematically mutated to reveal bias across protected attributes, i.e., race, gender and body. 
We use HInter~\citep{souani2025hinter} to perform single-attribute (atomic) and two-attribute (intersectional) mutations on 16,969 financial sentences from the Financial Sentiment Dataset (FinSen), producing \chan{125,161} original-mutant pairs.
\\
Aiming at (1) distinguishing bias-revealing inputs from others and (2) investigating results re-usability between model bias detection campaigns, we explore shared bias across models and quantify prediction shifts between original and mutated sentences for each model using Jensen-Shannon Distance (JSD) and Cosine Similarity.

Our results demonstrate that the studied models exhibit both atomic (0.58\% to 6.05\%) and intersectional (0.75\% to 5.97\%) bias, with varying magnitudes across attributes. 
These low bias ratios indicate that a very small proportion of inputs reveals bias, confirming that bias-revealing inputs are rare and most of the mutation and inference effort is in vain, further motivating the main goal of the study: reducing the cost of bias detection in language models.

More importantly, we show that a substantial portion of biased behavior can be uncovered by evaluating only a small fraction of selected input pairs.
In fact, the three lightweight models included in our study share over \chan{94\%} of their bias-revealing inputs, forming empirical evidence of the eventual cost-gain in simply reusing bias-revealing inputs across models, i.e., lightweight ones.
\\
Although different, bias-revealing inputs in generative models can be distinguished from the prediction probability shifts. For instance, we find that bias-revealing inputs in large models, i.e., FinMA, tend to yield more distant prediction probabilities between the original and mutated sentences.
In fact, prioritizing the test-inputs by decreasing Jensen–Shannon distance computed from lightweight model prediction scores, i.e., DistilRoBERTa, leads to an earlier bias discovery in other models, i.e., with just 20\% of the test-inputs, 73\% of FinMA’s bias is exposed, which is significantly larger than the $\approx20\%$ achieved by a random input-selection. 
These findings are encouraging and show a clear advantage in reusing prediction results between bias detection campaigns across models, particularly guiding bias detection in large (expensive) models using results from lightweight (cheap) ones.   

The contributions of this paper are threefold. First, we provide a comprehensive empirical analysis of bias in five financial language models using a large-scale dataset of real financial statements. Second, we identify shared patterns in bias-revealing inputs across models, offering insights into the transferability of bias detection efforts. Third, we demonstrate that bias detection can be significantly accelerated by prioritizing inputs based on model-derived 
features, reducing costs on large models. 
\\
We make our complete code base and setup instructions with additional details available~\footnote{\url{https://github.com/Firas-HadjKacem/FinancialBias.git}}
to enable reproducibility and support similar future research.


\ak{@ak we are using sentiment analysis as proxy to capture the prediction and perception of models}

\ak{may be add example of finbert with salary bias example}

\section{Related Work}
\label{sec:related_work}

\paragraph{Bias Detection in Language Models}
\label{subsec:bias_in_llms}
Social bias in language models often originates from imbalances and existing prejudices embedded in the training data, causing models to reflect and reproduce gender, racial, or cultural stereotypes~\citep{bender2021dangers,blodgett2020language}. 
Recent research has extensively studied such biases, consistently showing that they are pervasive across different architectures and tasks. Prior work has demonstrated that both generative~\citep{si2025detecting} and classification language models~\citep{kiritchenko2018examining} can encode and amplify societal stereotypes related to sensitive attributes.


Such bias in language models is typically studied through counterfactual (controlled-variation) testing~\citep{10.5555/3294996.3295162}, which consists of evaluating software output variation under the variation of controlled properties in the fed inputs~\citep{10.1145/2090236.2090255}. 
To this end, several datasets have been proposed, such as CrowS-Pairs, BBQ, and Winogender15,
which provide curated inputs of minimally different sentence pairs, with only demographic attribute changes~\citep{si2025detecting}. 
To not rely on existing datasets and favor realism in bias studies, recent works proposed automated input-generation approaches from existing real-world data. 
For instance, \citet{asyrofi2021biasfinder} proposed BiasFinder, a metamorphic testing approach that generates test cases by mutating sentences into input pairs that are expected to yield similar predictions, revealing a wider range of biases. 
Along the same lines of research, \citet{souani2025hinter} proposed HInter, an approach introducing one and two demographic attributes chnages per input, enabling the detection of not only atomic but also intersectional bias in the target models. 
In our work, we run HInter on the Financial Sentiment Dataset (FinSen)~\citep{FinSenDataset} to generate our test input pairs from real financial statements.   
As the detection relies on the variation of results, most studies on language models target classification tasks rather than generative ones. 
That is because model decision deviations are easier to distinguish between two different predicted labels compared to complex data, i.e., text.
Similarly, in our work, we target a classification task -- sentiment analysis -- as the proxy to study bias in financial language models. 

Unlike previous works, we investigate specific characteristics of bias-revealing input pairs, particularly common patterns within and across language models. 
Our study contributes to the language model bias detection research, providing empirical evidence that it can be significantly accelerated by prioritizing inputs based on model-derived 
features, reducing costs on large models. 
\paragraph{Financial Language Models}
\label{subsec:financial_language_models}
Over the past few years, a growing number of financial language models have been introduced, ranging from openly available models (e.g., FinBERT~\citep{araci2019finbert}, FinGPT~\citep{yang2023fingpt}, and FinMA~\citep{xie2023pixiu}) to proprietary solutions developed by private companies (e.g., BloombergGPT~\citep{wu2023bloomberggpt}).
These models have been trained or finetuned on top of backbone pre-trained general-purpose models, using large financial corpora.   
\\
Although very efficient in several financial downstream tasks~\cite{siala2026impact}, an evaluation of their fairness and biases, as well as a comprehensive assessment and comparison of different architecture types in this context is still lacking. 
\\
Our study is among the first to explicitly evaluate demographic fairness in financial language models. Given the increasing use of such models in high-stakes financial applications, these biases can potentially impact critical decisions in investment, lending, recruitment, etc. 
\section{Method}
\label{sec:method}
\begin{figure}[t]
\centering 
    \vspace{-0.8em}
\adjincludegraphics[width=0.9\textwidth, trim={{.0\width} {.0\width} {.0\width} {.0\width}} ,clip]
{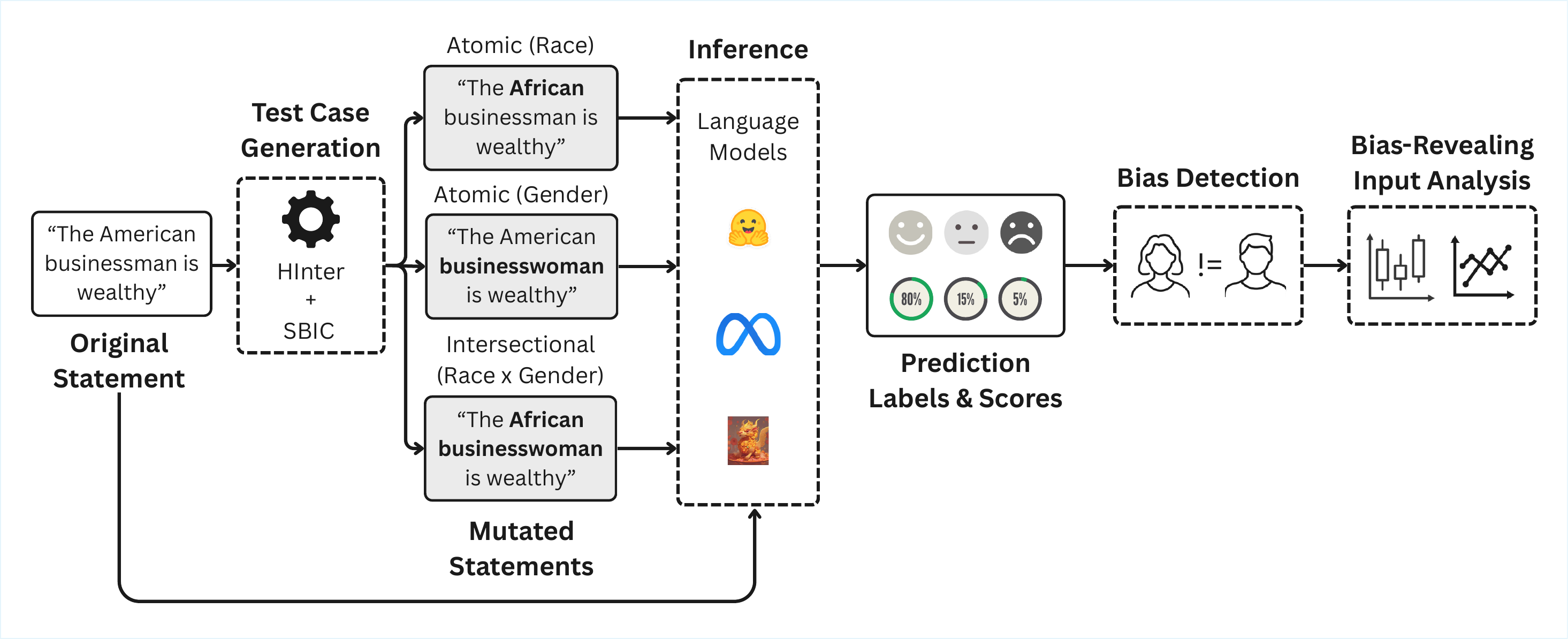}
\vspace{-1em}
\caption{The experimental workflow}
\label{fig:pipeline}
\vspace{-1em}
\end{figure}

Our experimental workflow consists of four main phases, as illustrated in Figure~\ref{fig:pipeline}: (1) the generation of bias test cases through input mutations, (2) sentiment prediction, (3) bias detection, and (4) cross-model bias-revealing inputs analysis. 
\subsection{Bias test-case Generation}
\label{subsec:bias_dictionary_test_case_generation}

This step consists of generating a dataset of test cases for the purpose of bias detection. To do so, we
use HInter~\citep{souani2025hinter}, a black-box metamorphic fuzzing approach,
which mutates input sentences by changing some properties (e.g., gender)
while keeping meaning and grammatical structure, generating several similar yet different versions of the original sentence. 
These replacements are employed using a bias dictionary derived from the Social Bias Inference Corpus (SBIC)~\citep{SBICdataset}, containing lists of terms and their substitutions for demographic categories. 
HInter focuses on three attribute axes commonly associated with bias: \textit{Gender}, \textit{Race} (the category grouping terms related to Ethnicity, National Origin, etc.), and \textit{Body} (the category grouping terms related to physical appearance). 
For example, the bias lexicon includes pairs of masculine$\leftrightarrow$feminine pronouns (e.g., ``he" $\leftrightarrow$ ``she"), gendered nouns (``businessman'' $\leftrightarrow$ ``businesswoman''), stereotypically gendered jobs (``CEO'' $\leftrightarrow$ ``Assistant''), nationality or ethnic descriptors (``American'' $\leftrightarrow$ ``Chinese''), racial terms (``Black'' $\leftrightarrow$ ``White''), and some body-related descriptors (``man'' $\leftrightarrow$ ``autistic man'').
\\
With this dictionary, we generated mutated versions of each input using two main mutation types:
\begin{itemize}
    \item \textbf{Atomic Mutations:} Changing one sensitive attribute at a time in the input sentence. \\ 
    For instance, applying a \textit{Gender} atomic mutation to the sentence “\textit{The CEO said \textbf{he} was confident about..}” produces the following mutated sentence \textit{“The CEO said \textbf{she} was confident about..”}, by changing \textbf{he} to \textbf{she}.
    \item \textbf{Intersectional Mutations:} Changing two sensitive attributes simultaneously in the input sentence. 
    For instance, applying a \textit{Gender \& Race} intersectional mutation to the sentence “\textbf{He} leads the \textbf{American} firm to success.” produces  “\textbf{She} leads the \textbf{Asian} firm to success.” by changing \textbf{He} and \textbf{American} to \textbf{She} and \textbf{Asian}, respectively. 
    \\
    These higher-order mutations are designed to reveal intersectional biases that might only emerge when two attribute changes co-occur.
\end{itemize}

\subsection{Model Inference and Bias Detection} 
\label{subsec:model_inference_bias_detection}



In this step, we ask the language models under test to predict the sentiment of each statement in our constructed dataset (original-mutant pairs).
We collect every inference result -- label and prediction probability of every class -- mapped with its corresponding input and prediction model, necessary for our bias detection and study analysis.
The sentiment analysis inference is operated differently depending on the target model's functioning:
\paragraph{Lightweight (Classifier) models}, i.e., FinBERT, DeBERTa-v3 (fine-tuned) and DistilRoBERTa (fine-tuned), offer the sentiment analysis task by design; taking as input the original sentence as-is and outputting its predicted sentiment together with the scores (logit values normalized by softmax) of each of the three possible labels (\textit{Negative, Positive and Neutral}) via its classification head. 
    
   \paragraph{Generative Language Models} , i.e., FinGPT and FinMA, 
    are instruction-based models, requiring a specific \textit{prompt} to process any task, i.e., the sentiment prediction.
    We follow the approach proposed by FinMA authors~\citep{xie2023pixiu} to perform classification tasks, adopting a zero-shot prompting approach to produce sentiment labels. 
    Specifically, we wrap each input \texttt{sentence} in a prompt asking for sentiment, as follows: 
    
    \begin{quote}
    Analyze the sentiment of this statement extracted from a financial news article.\\
    Provide your answer as either negative, positive, or neutral. Text: ``\textless \texttt{sentence}\textgreater''\\
    Answer:
    \end{quote}
 
    Beyond the functional design of the prompt, extra steps are required to extract labels and scores from the generated output text.
    In fact, models do not always follow the given instruction, e.g., not producing the sentiment directly after the prompt, potentially adding spaces or other misleading tokens before or in the middle of the label. 
    To tackle these issues, we restrict the vocabulary so that only token IDs that can begin a valid label variant are allowed (e.g, ``positive'', ``Positive'' with possible spacing or punctuation, etc.).   
    This ensures the model actually starts its answer with a sentiment label form. 
    At the same time, it enables the score and label extraction in the same forward pass, from the logits of the corresponding generated token. 
    \\
    Precisely, we parse the generated text to locate the first generated token corresponding to each of the sentiment labels, ignoring end-of-sequence or special tokens and including different label variants as spans (``Neg''+``ative''). 
    Next, we extract the model logits exactly at the found labels' locations.
   Finally, we attribute a score (probability) to each sentiment by softmaxing over the extracted logits, corresponding to each sentiment label.
   \\ 
   Whenever a label shows up in multiple tokenized forms, we combine them by using log-sum-exp before we apply softmax. This guarantees that variants of the sentiment labels (like ``negative'', `` Negative'') are fairly aggregated. 
   According to~\citet{goldberg2017nnlp}, ``Log-sum-exp'' is the numerically reliable way to add up probabilities in log space. With given log-scores $\{\ell_i\}$ for all possible tokenizations of a label, we compute $\displaystyle \log \!\sum_i e^{\ell_i}$, which equals the log of the sum of their prediction scores (probabilities). This enables us to compute the probability of the chosen label and compare it against other class logits in the same decoding step.
    By combining the logit extraction with the model’s native generational behavior, we obtain a hybrid logit extraction approach that avoids token misalignment issues and ensures we measure the model’s true confidence in the label it chose, exactly at the decision point. By ensuring consistency in predictions, we also guarantee comparable outputs to classifier models. 
    We employ greedy decoding to ensure determinism and reproducibility of our experiments. 
    
    
\subsection{Bias Detection}
\label{subsec:bias_detection}
As the only difference between each original-mutant pair is the demographic change, we expect a truly unbiased model to produce the same sentiment for both. 
Consequently, we consider any label change, e.g., from "Positive" to "Negative" sentiment, as a biased prediction and thus the input pair as bias-revealing. 
By comparing the prediction results collected in the previous step of each original-mutant pair, we obtain a set of bias-revealing inputs for each model, mutation type, and demographic property. 
Hence, the bias detection ratios of every model can be computed as the fraction of the number of bias-revealing inputs over the total number of inputs.

\subsection{Bias-Revealing Input Analysis}
\label{subsec:bias_metrics}
In this step, we investigate common trends and properties within bias-revealing inputs across models.
We start by studying the overlap and disjoint sets of bias-revealing inputs in the studied models. 
This is important, as large overlaps would enable direct reuse of inputs between bias-detection campaigns, saving inference effort and computational cost of finding bias inputs.

Moreover, we extend our study by investigating the overall decision shifts caused by mutations, beyond label flipping. 
To capture the decision shift, we compute the difference between the score vectors of every sentiment between the original-mutant pairs' predictions.
Using our prior definitions of probability distributions for sentences, we observe that $P(S_{\text{mutated}}) \neq P(S_{\text{orig}})$ whenever the input is altered, even if the predicted label remains the same. 
These shifts range from slight variations to ones large enough to cause a label flip. 
Our intuition is that such shifts can reveal subtle forms of bias not necessarily reflected in the output. 
More importantly, even when not revealing bias (no label prediction flip), such shifts could capture characteristics that differentiate bias-revealing inputs in the other models, enabling more cost-efficient cross-models guided bias-detection.

To measure the change in the output vectors indicative of eventual bias, we use two distance metrics:
 \paragraph{Jensen--Shannon Distance (JSD):} is the square root of Jensen--Shannon divergence. We chose it because it is symmetric and finite (bounded from 0 to 1), unlike Kullback--Leibler divergence, and retaining an information-theoretic meaning of how different two distributions are~\citep{Fuglede2004JSD}. 
    An input pair yields a $JSD=0$ if its vectors are identical $P(S_{\text{mutated}}) = P(S_{\text{orig}})$ and approaches 1 when they are maximally different. In the context of bias detection, JSD serves as a graded indicator of bias-induced change: a high JSD indicates a high probability shift caused by the demographic mutation.
    \paragraph{Cosine Similarity:} The metric has been widely used in measuring similarities between model captured context~\citep{salton1975vector}. 
    We use it to approximate the size of model prediction shifts by measuring the similarity between mutated and original prediction score vectors. This metric also takes floating numbers (0 to 1), where 1 indicates identical vectors and 0 completely different ones. 

\subsection{Cross-model Guided Bias Detection}
\label{subsec:bias_simulation}
\ak{talk about it in the analysis and describe it more in the experimental protocol...}
To investigate the practical usefulness of the observed common properties and trends between models, we propose a cross-model guided bias detection approach. 
Unlike conventional random approaches, we prioritize the set of input test pairs based on their prediction results from other models.
To evaluate this approach's effectiveness and cost-efficiency, we compare it with a random one, collecting bias-detection rates at every effort step in terms of input-pair inference.

Although this approach could benefit any bias detection campaign, we focus particularly on assessing the advantage of guiding bias detection in large, expensive models using lightweight, cheaper ones. 
We refer to the classifiers (FinBERT, DeBERTa-v3 (fine-tuned), DistilRoBERTa (fine-tuned)) as the lightweight, smaller models, used as reference, and refer to the generative LLMs (FinMA, FinGPT) as the larger models, used as targets.

On the six combinations of lightweight-large models, we run (100 times) bias detection campaigns applying each of the following input prioritization strategies:
\\
    - Random input ordering as baseline, \\
    - Ordering inputs by decreasing JSD in the reference model’s predictions,\\
    - Ordering inputs by increasing cosine similarity in the reference model’s predictions,\\
    - Prioritizing inputs where the reference model flagged a bias 







\section{Experiments}
\label{sec:experiments}

\begin{table*}[t]
\centering
\vspace{-10pt}
\caption{\small Models bias rates on FinSen. Total tested pairs, per attribute: Body = 22,972; Gender = 105,962; Race = 121,490, overall = 125,161.}
\ak{for every one of these percentages, what is the nominator and denominator}
\label{tab:finsen-bias-summary}
\small
\setlength{\tabcolsep}{3.0pt}
\scalebox{0.9}{
\begin{tabular}{lccccccccc}
\toprule
\textbf{Model} &
\makecell{\textbf{Atomic}\\\textbf{(Body)}} &
\makecell{\textbf{Inter.}\\\textbf{(Body)}} &
\makecell{\textbf{Atomic}\\\textbf{(Gender)}} &
\makecell{\textbf{Inter.}\\\textbf{(Gender)}} &
\makecell{\textbf{Atomic}\\\textbf{(Race)}} &
\makecell{\textbf{Inter.}\\\textbf{(Race)}} &
\makecell{\textbf{Total}\\\textbf{(Atomic)}} &
\makecell{\textbf{Total}\\\textbf{(Inter.)}} &
\makecell{\textbf{Total Hidden}\\\textbf{(Inter.)}} \\
\midrule
FinMA         & 9.23\% & 7.48\% & 2.77\% & 2.25\% & 3.25\% & 3.29\% & 3.99\% & 3.23\% & 4.05\% \\
FinGPT        & 5.39\% & 2.77\% & 6.10\% & 6.55\% & 6.13\% & 6.07\% & 6.05\% & 5.97\% & 31.29\% \\
FinBERT       & 1.89\% & 1.88\% & 0.69\% & 0.88\% & 0.25\% & 0.41\% & 0.58\% & 0.75\% & 30.34\% \\
DeBERTa-v3    & 1.69\% & 1.67\% & 0.70\% & 0.89\% & 0.30\% & 0.46\% & 0.60\% & 0.75\% & 29.95\% \\
DistilRoBERTa & 1.69\% & 1.67\% & 0.70\% & 0.89\% & 0.30\% & 0.46\% & 0.60\% & 0.75\% & 29.95\% \\
\bottomrule
\end{tabular}
}
\vspace{-10pt}
\end{table*}

\subsection{Experimental Setup}
\label{subsec:experimental_setup}

\paragraph{Dataset:}
To conduct this study, we use the US subset of the Financial Sentiment Dataset (FinSen) \citep{FinSenDataset}, featuring 16,969 financial sentences. FinSen provides a realistic sample of finance-related text spanning 15 years, extracted from financial news articles, headlines, and financial statements.
Sentence length varies from small (42 words) to long (186 words) statements.
We mutate this dataset as described in Section~\ref{subsec:bias_dictionary_test_case_generation} to construct 125,161 original-mutated bias test cases. 
We keep information relative to the attribute(s) changed in each pair, effectively partitioning the augmented corpora by mutation type.
\paragraph{Financial Language Models:}
We conduct our study on 
five financial language models including 
2 multi-purpose generative LLMs and 3 encoder-based classifiers, obtained through fine-tuning on financial corpora:
- FinMA: a 7B-parameter LLaMA2-based generative financial LLM from the PIXIU study\citep{xie2023pixiu}. It was extensively trained on financial data and was specifically fine-tuned for sentiment classification to generate textual sentiment labels when prompted.
    \\
    - FinGPT: also a 7B-parameter LLaMA2-based generative LLM, that was open-sourced by~\citet{yang2023fingpt}. It was adapted for financial sentiment generation through additional financial corpus pre-training and lightweight fine-tuning via LoRA.
    \\
    - FinBERT: a classifier, classic BERT-based (110M parameters) financial sentiment model. It has been further pre-trained on financial texts, then fine-tuned on financial sentiment classification (using datasets like Financial PhraseBank and FiQA) \citep{araci2019finbert}.
    \\
    - DeBERTa-v3 (fine-tuned, ~\citep{romero2024deberta}): a classifier DeBERTa-v3 base model~\citep{He2021DeBERTa}, fine-tuned on financial news sentiment ($\approx$44M parameters).
    \\
    - DistilRoBERTa (fine-tuned,~\citep{romero2024distilroberta}): a classifier, distilled RoBERTa-base model ~\citep{Liu2019RoBERTa}, fine-tuned on financial news sentiment. It has $\approx$82M parameters.
\\
The links to the model Hugging Face IDs are listed in Table~\ref{financial_language_models} in the Appendix Section~\ref{appendix_models}.
\subsection{Results}
\label{subsec:results}
\subsubsection{Bias Detection Effectiveness in Financial Language Models}
\label{subsubsec:bias_detection}
The five studied models showed biased behaviours with different magnitudes for the different mutations (atomic and intersectional) and wrt the three controlled properties. 
The percentage of bias-detecting pairs from our dataset is reported in Table~\ref{tab:finsen-bias-summary}, by model and controlled property. 
Perhaps surprisingly, we notice that simple and lightweight models, i.e., FinBERT, DeBERTa-v3, and DistilRoBERTa, exhibit lower overall atomic~($\approx0.6\%$) and intersectional~($\approx0.75\%$) bias ratios compared to larger ones, i.e., FinMA and FinGPT, scoring respectively 3.99\% and 6.05\% as atomic and 3.23\% and 5.97\% as intersectional bias. 
This encourages the use of small classification language models over larger ones as a bias mitigation measure.
The analysis of intersectional bias shows that a large proportion of it is \textit{hidden (not discovered by single property mutation)}; $\approx30\%$ of FinBERT, DeBERTa-v3 and DistilRoBERT, $\approx31\%$ of FinGPT, and $\approx4\%$ of FinMA, highlighting the impact of going beyond single-attribute bias detection by using higher-order mutations. 

\ak{The results show that a very small proportion of inputs (less than 6\% of the studied dataset) is revealing bias. 
These findings confirm that the bias-revealing inputs are rare and most if the muation and inference effort is vain, motivating further the main goal of the study; reducing the cost of bias detection in language models.}

\subsubsection{Shared Bias Across Models}
\label{subsubsec:bias_overlaps}
\noindent
\begin{minipage}{0.53\linewidth}

The Venn diagram of Figure~\ref{fig:model-overlap} depicts the overlap and disjoint sets of bias-revealing input-pairs in the five studied models.
Although common bias triggers exist, we notice that there is no universal set of common bias-revealing inputs across models. 
However, we notice a clear overlap between models of similar architectural families, particularly between the three studied classifiers, with over 94\% of their bias being revealed by the same inputs and a full overlap between DeBERTa-v3 and DistilRoBERTa sets. 
\\
These results are encouraging and endorse the re-usability of inputs between lightweight models, as one can uncover bias-revealing inputs in a model without running a full bias-detection campaign. 
However, despite their similar training and capacity, generative models share only a small set of biased inputs (9). Together with the negligible overlap with the lightweight models, this encourages further investigation into common properties between bias-revealing inputs across models.


\end{minipage}\hfill
\begin{minipage}{0.45\linewidth}
\centering
\adjincludegraphics[width=\linewidth, trim={{.09\width} {.08\width} {.04\width} {.05\width}} ,clip]
{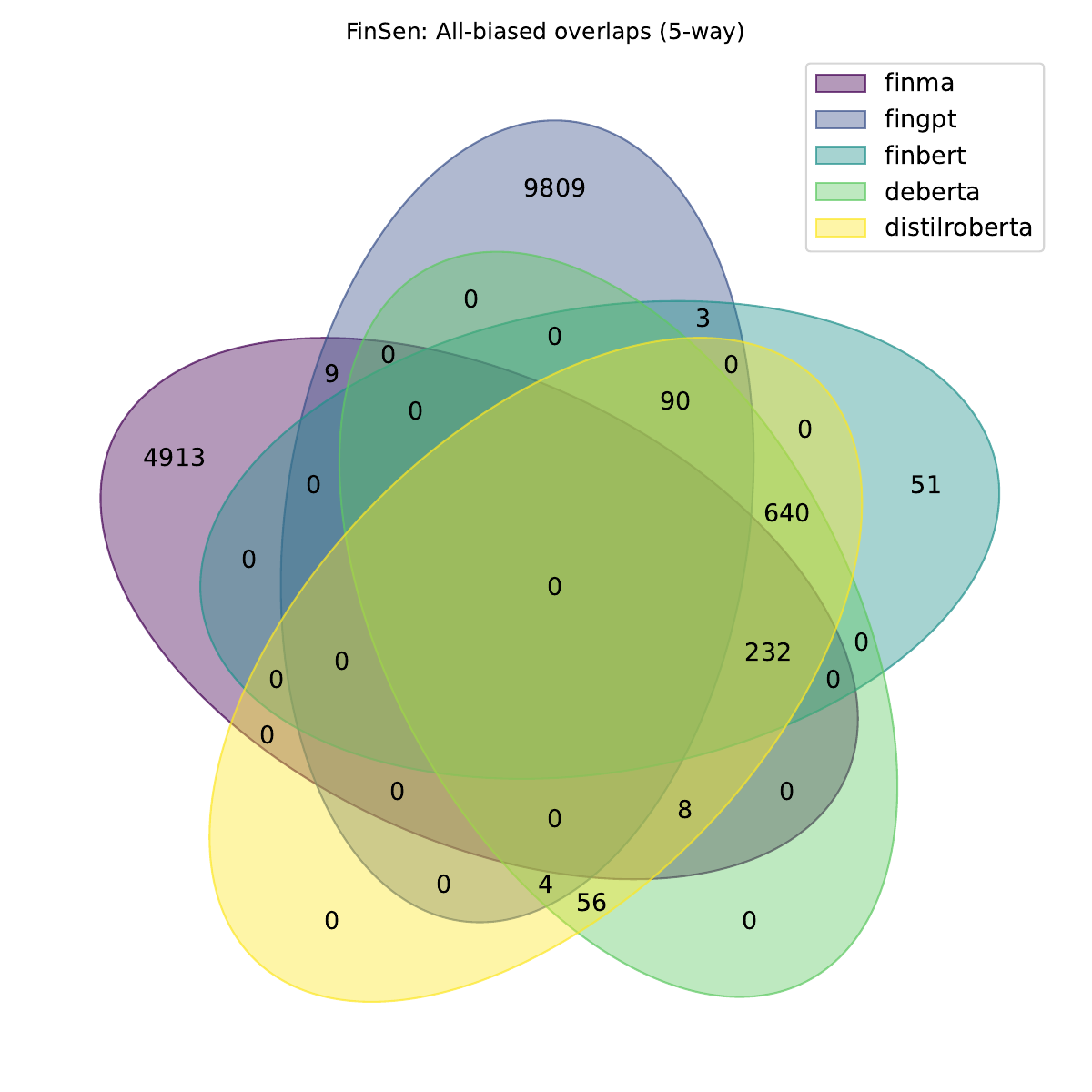}
\captionof{figure}{Bias overlaps across models.}
\label{fig:model-overlap}
\end{minipage}

\subsubsection{Cross-Models Bias Revealing Input-Pairs Analysis}
\label{subsubsec:bias_quantification}

\ak{larger fonts on figures. save as pdf. write model names as in paper: capital letters... no need to try to change the venn diagram, it's fine.}

Aiming at reducing model bias detection costs by distinguishing its bias-revealing input pairs prior to inference, we extend our study by investigating the magnitude of the decision shifts caused by the mutation in other models. 
More precisely, for a model~\textit{M1} we compute the cosine and JSD distances between the original sentence and its mutated counterpart, each represented by its prediction scores (probabilities) obtained from a second model~\textit{M2}.
For instance, we illustrate in Figure~\ref{fig:jsdfinmabert} the difference between FinMA's~\textit{(M1)} bias-revealing pairs from non-bias-revealing ones, in terms of JSD computed on the predictions of each of the lightweight models~\textit{(M2)}.
The box plots indicate that bias-revealing pairs tend to score more distant predictions than the non-revealing ones; with median JSD of \chan{$\approx0.031$} and \chan{$\approx0.023$} compared to \chan{$\approx0.003$} and \chan{$\approx0.002$} for respectively intersectional and atomic test cases. 
\\
We observe similar distinguishing trends with different magnitudes when targeting the other models or computing the cosine similarity instead of the JSD. 


\begin{figure}[t]
\label{fig:jsdfinmabert}
\centering
\vspace{-1em}
\begin{subfigure}{0.4\textwidth}
\adjincludegraphics[width=\textwidth, trim={{.008\width} {.01\width} {0.33\width} {.07\width}} ,clip]{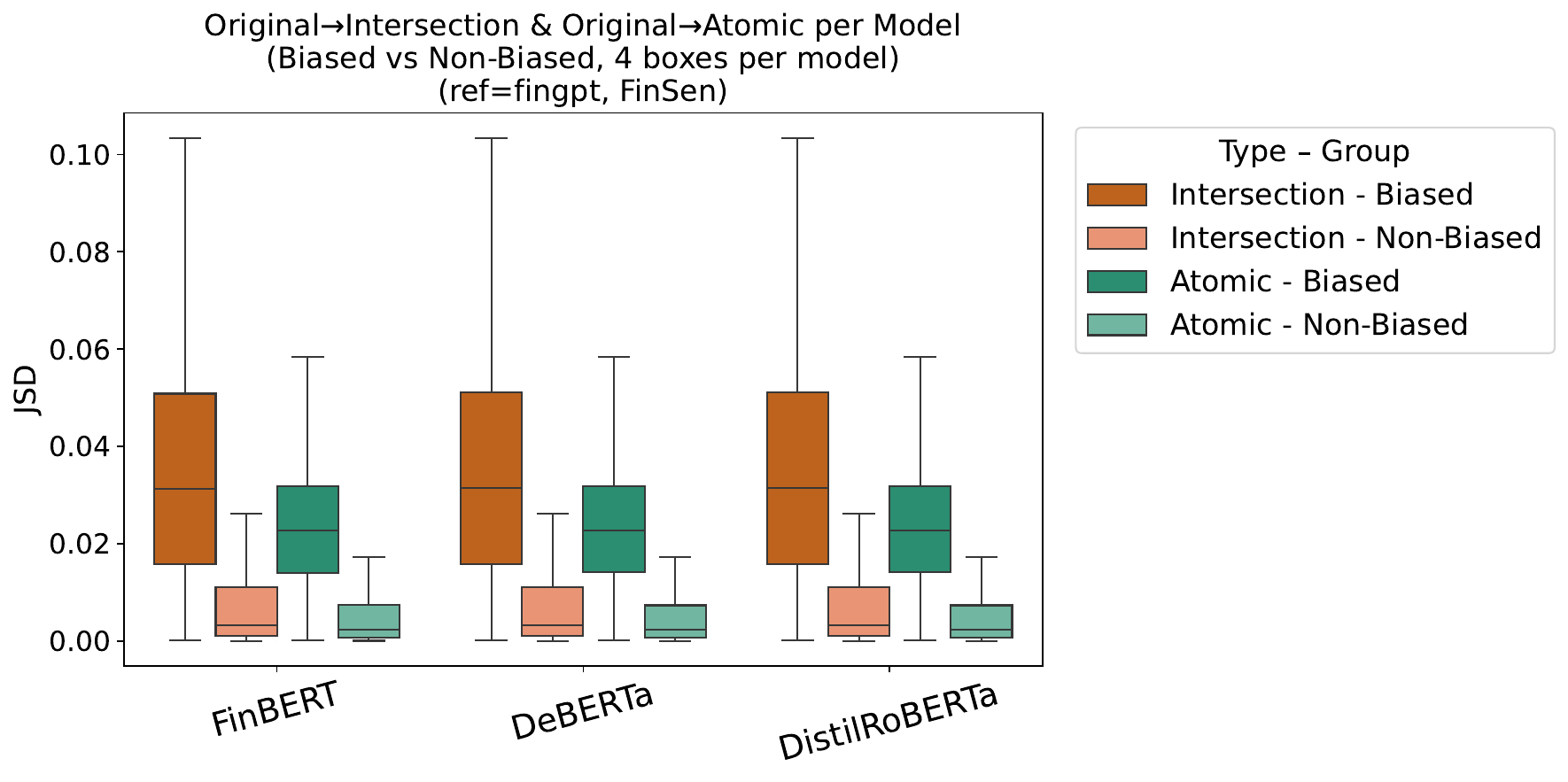}
\caption{{\small Lightweight models prediction shifts.}}
\label{fig:jsdfinmabert}
\end{subfigure}
\begin{subfigure}{0.58\textwidth}
\centering
\adjincludegraphics[width=\textwidth, trim={{0.008\width} {.01\width} {.008\width} {.07\width}} ,clip]{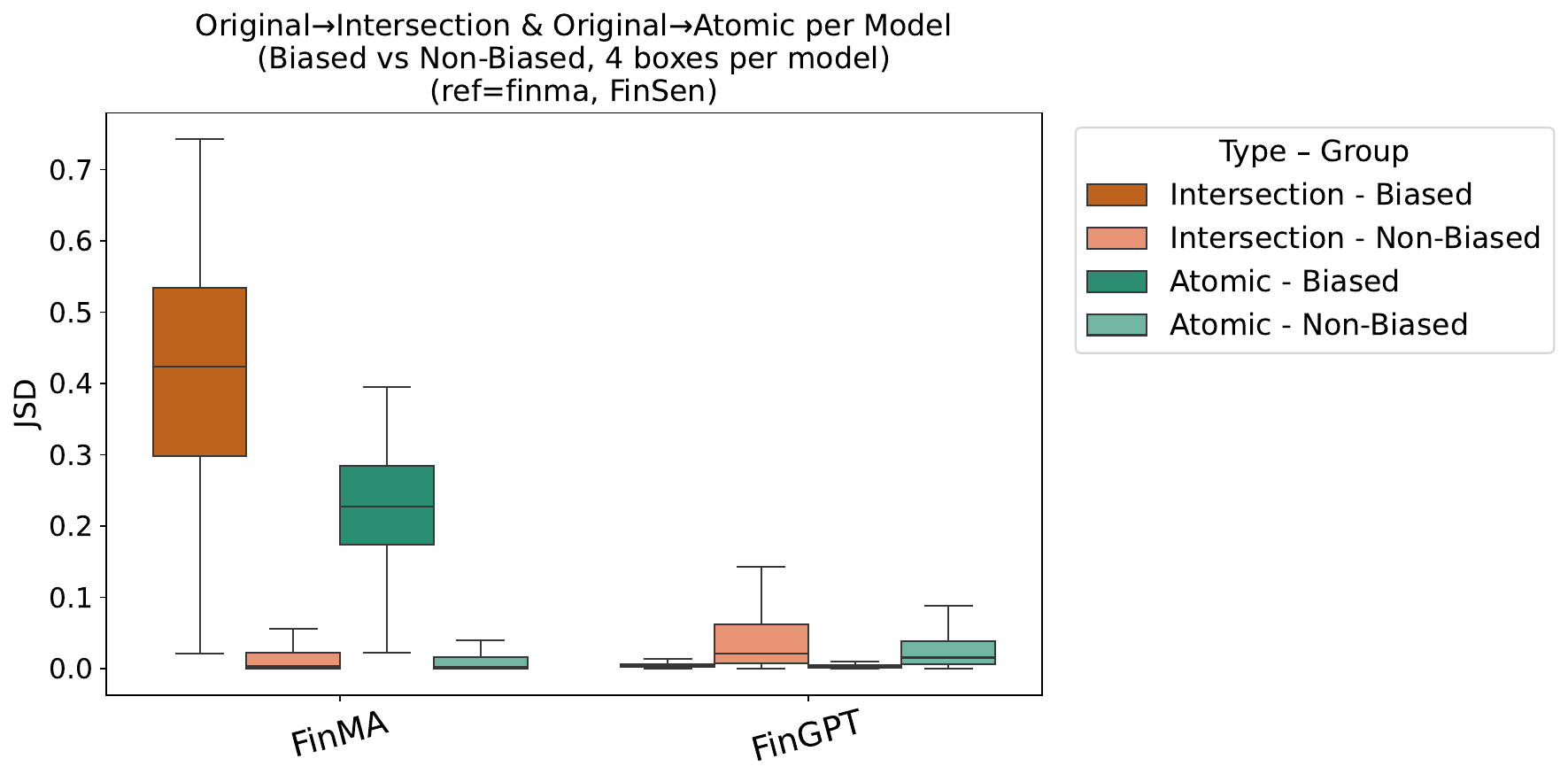}
\caption{{\small Large models prediction shifts.}}
\label{fig:jsdfinmallama}
\end{subfigure}
\caption{\small Comparison of FinMA bias-revealing and non-bias-revealing inputs, based on the JSD distance between original and mutated statements computed on their prediction probabilities, obtained from lightweight and large models. 
}

\vspace{-1em}
\label{fig:bp_jsd_finma}
\end{figure}

\noindent
\begin{minipage}{0.4\linewidth}
Very low \chan{Wilcoxon} p-values (below \chan{$10^{-32}$}) and very far from 0.5 \VDA values (over \chan{0.35} difference, as illustrated in Table~\ref{tab:vda_bp_jsd_finma}) indicate that the observed differences are significant and not happening by chance, confirming that in the majority of the cases, bias-revealing inputs can be effectively detected based on another model's prediction shifts. 

\end{minipage}\hfill
\begin{minipage}{0.58\linewidth}
\centering

    \vspace{-1.5em}
   \captionof{table}{ \small Vargha and Delaney \VDA effect size values of prediction shifts difference between FinMA bias-revealing and non bias-revealing inputs (computed on values of Figure\ref{fig:bp_jsd_finma}).
}
    \scalebox{0.7}{
    \begin{tabular}{ccclll}\toprule
         \VDA&  FinBERT& DeBERTa & DistilRoBERTa&FinMA &FinGPT\\\midrule
         Atomic& 0,88 & 0,88 &0,88 & 0,99 & 0,16\\
         Intersectional& 0,85 & 0,85 &0.85 & 0,99 & 0,18\\ \bottomrule
    \end{tabular}
    }
    \vspace{-1em}
    \label{tab:vda_bp_jsd_finma}
\end{minipage}

\ak{wicoxon ref}

\subsubsection{Cost-efficiency of cross-model guided bias detection}
\label{subsubsec:bias_acceleration}

\ak{adjust table}

\begin{table}[]
\caption{\small Comparison of the efficiency and cost efficiency of a guided input selection with a conventional (random) one, in detecting large model bias based on DistilRoBERTa predictions. 
We note respectively BD\%, $\Delta$, p-val, and \VDA, the bias detection ratio, the BD\% difference with random, \chan{Wilcoxon} p-values, and \VDA values.}
\centering
\setlength{\tabcolsep}{3pt}
\scalebox{0.68}{
\begin{tabular}{|l|llll|llll|llll|llll|}
\toprule
    \textbf{Model / }
    & \multicolumn{4}{c|}{\textbf{20\%}} 
    & \multicolumn{4}{c|}{\textbf{40\%}} 
    & \multicolumn{4}{c|}{\textbf{60\%}}
    & \multicolumn{4}{c|}{\textbf{80\%}}
    \\  
    \textbf{Effort}
    & \multicolumn{1}{l}{BD\%} & $\Delta$ & p-val & \VDA
    & \multicolumn{1}{l}{BD\%} & $\Delta$ & p-val & \multicolumn{1}{l|}{\VDA}  
    & \multicolumn{1}{l}{BD\%} & $\Delta$ & p-val & \multicolumn{1}{l|}{\VDA} 
    & \multicolumn{1}{l}{BD\%} & $\Delta$ & p-val & \multicolumn{1}{l|}{\VDA} 
    \\ \midrule
    \multicolumn{17}{|l|}{\textbf{Prioritization ranking DistilRoBERTa bias revealing inputs first.}}
    \\ \midrule
    \textbf{FinMA}  
        & \multicolumn{1}{l}{23,12}   
        & \textbf{+3,12} 
        & \multicolumn{1}{l}{$\mathcal{O}(10^{-18})$}   
        & \multicolumn{1}{l|}{1} 
        & \multicolumn{1}{l}{42,35}   
        & \textbf{+2,35} 
        & \multicolumn{1}{l}{$\mathcal{O}(10^{-18})$}   
        & \multicolumn{1}{l|}{0,996}
        & \multicolumn{1}{l}{61,53}   
        & \textbf{+1,53} 
        & \multicolumn{1}{l}{$\mathcal{O}(10^{-17})$}   
        & \multicolumn{1}{l|}{0,936} 
        & \multicolumn{1}{l}{80,79}  
        & \multicolumn{1}{l}{\textbf{+0,79}}
        & \multicolumn{1}{l}{$\mathcal{O}(10^{-13})$}   
        & \multicolumn{1}{l|}{0,835}
        \\ 
    \textbf{FinGPT}                       
        & \multicolumn{1}{l}{20,07}  
        & \textbf{+0,07} 
        & \multicolumn{1}{l}{$\mathcal{O}(10^{-2})$}   
        & \multicolumn{1}{l|}{0,58} 
        & \multicolumn{1}{l}{40,02}   
        & \multicolumn{1}{l}{\textbf{+0,02}}
        & \multicolumn{1}{l}{$\mathcal{O}(10^{-1})$}   
        & \multicolumn{1}{l|}{0,52}
        & \multicolumn{1}{l}{60,03}   
        & \textbf{+0,03}
        & \multicolumn{1}{l}{$\mathcal{O}(10^{-1})$}   
        & \multicolumn{1}{l|}{0,52} 
        & \multicolumn{1}{l}{80,04}   
        & \multicolumn{1}{l}{\textbf{+0,04}}
        & \multicolumn{1}{l}{$\mathcal{O}(10^{-1})$}   
        & \multicolumn{1}{l|}{0,52}
        \\ 
    \midrule
    \multicolumn{17}{|l|}{\textbf{Prioritization of inputs by decreasing JSD of their DistilRoBERTa prediction shifts.}}
    \\ 
    \midrule
    \textbf{FinMA}  
        & \multicolumn{1}{l}{73,01}   
        & \textbf{+53,01} 
        & \multicolumn{1}{l}{$\mathcal{O}(10^{-18})$}   
        & \multicolumn{1}{l|}{1} 
        & \multicolumn{1}{l}{89,64}   
        & \multicolumn{1}{l}{\textbf{+49,64}}
        & \multicolumn{1}{l}{$\mathcal{O}(10^{-18})$}   
        & \multicolumn{1}{l|}{1}
        & \multicolumn{1}{l}{95,49}   
        & \textbf{+35,49} 
        & \multicolumn{1}{l}{$\mathcal{O}(10^{-18})$}   
        & \multicolumn{1}{l|}{1} 
        & \multicolumn{1}{l}{97,5}  
        & \multicolumn{1}{l}{\textbf{+17,5}}
        & \multicolumn{1}{l}{$\mathcal{O}(10^{-18})$}   
        & \multicolumn{1}{l|}{1}
        \\
    \textbf{FinGPT}                       
        & \multicolumn{1}{l}{7,61}   
        & \textbf{-12,39} 
        & \multicolumn{1}{l}{1}   
        & \multicolumn{1}{l|}{0} 
        & \multicolumn{1}{l}{34,14}   
        & \multicolumn{1}{l}{\textbf{-5,86}}
        & \multicolumn{1}{l}{1}   
        & \multicolumn{1}{l|}{0}
        & \multicolumn{1}{l}{61,52}   
        & \textbf{+1,52}
        & \multicolumn{1}{l}{$\mathcal{O}(10^{-18})$}   
        & \multicolumn{1}{l|}{1} 
        & \multicolumn{1}{l}{82,76}   
        & \multicolumn{1}{l}{\textbf{+2,76}}
        & \multicolumn{1}{l}{$\mathcal{O}(10^{-18})$}   
        & \multicolumn{1}{l|}{1}
        \\ 
\bottomrule
\end{tabular}}
\label{tab:RQ3_a12}
\vspace{-1.0em}
\end{table}

To evaluate the practical usefulness of the observed overlaps and differences in prediction shifts (findings of Sections~\ref{subsubsec:bias_overlaps} and~\ref{subsubsec:bias_quantification}~) between input pairs, we compare the efficiency and cost-efficiency of a guided selection with a conventional (random) one.
\\ 
As lightweight models share most of their bias-revealing inputs (findings of Section~\ref{subsubsec:bias_overlaps}), guiding any of them by another leads to near full bias detection at very low cost (95\%+ at $<20\%$ of inputs).

Table~\ref{tab:RQ3_a12} reports the large models' detected bias ratio by effort spent in terms of input pairs' predictions, when guided by those of a lightweight model, i.e., DistilRoBERTa, compared to a random approach.
\\
The positive $\Delta$ values scored for both FinMA and FinGPT depict an advantage in prioritizing DistilRoBERTa bias-revealing inputs over random, at low and high cost. 
These differences are relatively small, which can be explained by the small overlap between bias revealing input sets of the considered models, as seen in Section~\ref{subsubsec:bias_overlaps}.
When prioritizing input pairs with larger JSD computed on DistilRoBERTa prediction probabilities, improvements over random ordering remain negligible for FinGPT and are only observable at higher effort.
However, when applied to FinMA, this strategy shows a clear advantage, uncovering 73,01\% and 89,64\% of FinMA bias at only 20\% and 40\% of the effort.  
Conducted statistical tests and size-effect measures validate the significant advantage of the proposed input-prioritization strategy over random, with p-values in the order of {$10^{-18}$} and \VDA values of {$\approx1$}.

\section{Conclusion}
\label{sec:conclusion}
This paper presented a large-scale study of demographic bias in financial language models, showing that large and lightweight models exhibit bias under both atomic and intersectional settings. 
This work represents the first comprehensive investigation of bias detection in financial language models. We further demonstrated that bias detection can be significantly accelerated by exploiting shared patterns in bias-revealing inputs across models, enabling a large fraction of biased behaviors to be uncovered using only a small subset of test inputs. In doing so, we provide the first empirical evidence supporting cross-model–guided bias detection as a promising direction for reducing the cost in bias auditing and any downstream bias-related tasks, i.e., mitigation, by providing practitioners with essential bias-revealing inputs at significantly lower costs.
Although obtained in the financial domain, our findings may generalize to other language models and application domains, which we leave for future work.

\section*{Acknowledgment}
This research was funded in whole, or in part, by the Luxembourg National Research Fund (FNR), grant reference NCER22/IS/16570468/NCER-FT.

\bibliography{iclr2026_conference}
\bibliographystyle{iclr2026_conference}

\newpage

\appendix
\section{Appendix}
\label{appendix}
\subsection{Financial Language Models}
\label{appendix_models}



\begin{table}[H]
\caption{Links to the studied Financial language models in HugginFace}
\label{financial_language_models}
\begin{center}
\setlength{\tabcolsep}{4pt}
\renewcommand{\arraystretch}{1.1}

\begin{tabularx}{\linewidth}{@{} l >{\raggedright\arraybackslash}X @{}}
\toprule
\multicolumn{1}{c}{\bf Model} & \multicolumn{1}{c}{\bf HuggingFace ID}
\\ \hline 
FinMA         & \url{https://huggingface.co/ChanceFocus/finma-7b-full} \\
FinGPT        & \url{https://huggingface.co/oliverwang15/FinGPT_v32_Llama2_Sentiment_Instruction_LoRA_FT} \\
FinBERT       & \url{https://huggingface.co/ProsusAI/finbert} \\
DeBERTa-v3    & \url{https://huggingface.co/mrm8488/deberta-v3-ft-financial-news-sentiment-analysis} \\
DistilRoBERTa & \url{https://huggingface.co/mrm8488/distilroberta-finetuned-financial-news-sentiment-analysis} \\
\bottomrule 
\end{tabularx}
\end{center}
\end{table}
\subsection{Implementation details}
\label{implementation}
We implemented our
experimental workflow as described earlier in Section~\ref{sec:method} as python scripts, relying on well-established publicly available libraries, namely: 
\\
- transformers for model loading, tokenization, and inference, 
\\
- bitsandbytes for model quantization, 
\\
- Pytorch for inference and on-GPU execution operations, 
\\
- LogitsProcessor for vocabulary masking and constrained generation.
\\
- numpy and pandas for several vector operations.
\\
Other libraries and tools were also used throughout the codebase. All of these were important to load the financial models, generate predictions, obtain bias results, and quantify bias via JSD and Cosine similarity.\\ 
We fixed random seeds for any stochastic operations (e.g., text generation 
) to ensure reproducibility of results.
\\
The implementation of our approach and the reproduction package are available at \url{https://github.com/biasfinllms/iclr-finai-submission}

\subsection{Compute infrastructure and resources}
\label{compute_infrastructure}
    \paragraph{Test Case Generation:} We performed the mutation generation and dataset preprocessing on a standard workstation (Intel Core i7-1165G7 @ 2.8GHz CPU, 16GB RAM) without GPU acceleration, taking only a few hours to complete.
     \paragraph{Model inference and analysis:} To run the sentiment models on the 125k+ test cases and perform bias analysis, we utilized GPU-enabled compute environments. Initial prototyping and spot-checking of model outputs were conducted on a single NVIDIA T4 GPU (via Google Colab’s free tier). However, the full-batch inference over all mutated pairs 
     was
     more computationally intensive. To run these experiments, we utilized a high-performance server equipped with 3 × NVIDIA RTX A6000 GPUs (each with 49 GB of VRAM) and 504 GB of system RAM.  

\end{document}